\documentclass[conference]{IEEEtran}
\IEEEoverridecommandlockouts
\usepackage{cite}
\usepackage{amsmath,amssymb,amsfonts}
\usepackage{algorithmic}
\usepackage{graphicx}
\usepackage{textcomp}
\usepackage{xcolor}
\usepackage{multirow}
 
\usepackage{svg}
\def\BibTeX{{\rm B\kern-.05em{\sc i\kern-.025em b}\kern-.08em
    T\kern-.1667em\lower.7ex\hbox{E}\kern-.125emX}}

\begin{document}
\title{Agile gesture recognition for capacitive sensing devices: adapting on-the-job
\thanks{The project was supported by the Innovate UK KTP grant (12250) and by the UKRI Turing AI Acceleration Fellowship EP/V025295/1, EP/V025295/2.}}

\author{

\IEEEauthorblockN{1\textsuperscript{st} Ying Liu}
\IEEEauthorblockA{  \textit{University of Leicester}\\
                    Leicester, U.K. \\
                    0000-0002-2537-5429
                }

\and
\IEEEauthorblockN{2\textsuperscript{nd} Liucheng Guo}
\IEEEauthorblockA{\textit{Tangi0 Ltd.} \\
                    London, U.K. \\
                    0009-0008-9419-4075}
\and
\IEEEauthorblockN{3\textsuperscript{rd} Valeri A. Makarov}
\IEEEauthorblockA{\textit{Universidad Complutense de Madrid} \\
                    Madrid, Spain \\
                    0000-0001-8789-7532}
\and
\IEEEauthorblockN{4\textsuperscript{th} Yuxiang Huang}
\IEEEauthorblockA{\textit{University of Leicester}\\
                    Leicester, U.K. \\
                    yh351@leicester.ac.uk}

\and
\IEEEauthorblockN{5\textsuperscript{th} Alexander Gorban}
\IEEEauthorblockA{ \textit{University of Leicester}\\
                    Leicester, U.K. \\
                    0000-0001-6224-1430}                    
\and
\IEEEauthorblockN{6\textsuperscript{th} Evgeny Mirkes}
\IEEEauthorblockA{ \textit{University of Leicester}\\
                    Leicester, U.K. \\
                    0000-0003-1474-1734}
                    
\and
\IEEEauthorblockN{7\textsuperscript{th} Ivan Y. Tyukin}
\IEEEauthorblockA{  \textit{King's College London} \\
                    London, U.K. \\
                    0000-0002-7359-7966}

}
                    
\maketitle

\begin{abstract}
Automated hand gesture recognition has been a focus of the AI community for decades. Traditionally, work in this domain revolved largely around scenarios assuming the availability of the flow of images of the operator's/user's hands. This has partly been due to the prevalence of camera-based devices and the wide availability of image data. However, there is growing demand for gesture recognition technology that can be implemented on low-power devices using limited sensor data instead of high-dimensional inputs like hand images.

In this work, we demonstrate a hand gesture recognition system and method that uses signals from  capacitive sensors embedded into the \textit{etee} hand controller. The controller generates real-time signals from each of the wearer's five fingers. We use a machine learning technique to analyse the time-series signals and identify three features that can represent 5 fingers within 500 ms. The analysis is composed of a two-stage training strategy, including dimension reduction through principal component analysis and classification with K-nearest neighbour. Remarkably, we found that this combination showed a level of performance which was comparable to more advanced methods such as supervised variational autoencoder. The base system can also be equipped with the capability to learn from occasional errors by providing it with an additional adaptive error correction mechanism.  The results showed that the error corrector improve the classification performance in the base system without compromising its performance. The system requires no more than 1 ms of computing time per input sample, and is smaller than deep neural networks, demonstrating the feasibility of agile gesture recognition systems based on this technology. 
\end{abstract}


\section{Introduction}

Hand gesture recognition algorithms have developed intensively in recent years due to the advancements in technology and the increased availability of personal camera devices \cite{Oudah2020review}. 
There are two main approaches for recognising hand gestures: 1) computer vision-based systems, which use advanced algorithms to detect hand gestures from image data; or 2) hardware-based embedded systems, which measure signals from muscle movement and classify them using software. 
Hardware-based embedded systems have the potential to quickly measure signals induced by movements of limbs or muscles directly. This has an advantage over the alternatives that rely upon the interpretation of high dimensional image data. The speed is important for a broad range of relevant scenarios including human-computer interaction, human behaviour analysis, and accessibility solutions for people with movement disorders \cite{Benalcazar2017Myo, Wang2020GR, Moin2020wearable, Oweis2011EEG}. On top of that, they vastly reduce the risks of accidental or adversarial leakage of identifiable personal information. This is achieved by avoiding the need to capture video and/or photographic imagery as a part of the gesture acquisition process.
These hardware-based systems rely on various types of signals for the gesture recognition, including a combination of wire and spring to measure joint angle \cite{Park2015,Li2011}, hetero-core flexion sensors \cite{Nishiyama2009}, inertial measurement unit\cite{Xing2017IMU}, piezoresistive sensor \cite{Sundaram2019},  capacitive sensor and electromyography \cite{Benalcazar2017Myo}. 

Regardless, however, of how the gesture signals are measured,  the second major task is to recognise or classify the information contained in the physical signals. Currently, the most common hand gesture recognition algorithms use neural networks (NNs). This is because NNs are effective for classifying high dimensional data, e.g. image, which are the primary analysis method for computer-vision based gesture recognition \cite{Oudah2020review}. 
While NNs have proven effective for gesture recognition thus far, state-of-the-art models typically require very large datasets \cite{Bambach2015ICCV} of pre-labelled data. In addition, real-time inference with NN models may require levels of computing resources which are not available or feasible for low-power embedded systems (less than 1 W). The majority of NN model can only be implemented on edge devices which requires the power consumption exceeded 5 W \cite{deepedgebench2021}. These power requirements present a challenge limiting and hindering the scope of applications of hardware-based embedded systems and systems with cheap capacitive sensors in particular. Our goal is to create an agile gesture recognition system that can operate on a low-power edge device for live prediction. Additionally, we aim to incorporate a feature that allows the system to adapt to a user's customised gesture patterns and correct itself accordingly.

In this work, we propose an approach which can be used to address these purposes. The approach is based on a combination of agile and fast classical recognition model equipped with an "adaptation add-on" capable to fine-tune the model to the end-user. To demonstrate the capability, we used the hand controller \textit{etee} for hand movement signal recording through Bluetooth \cite{etee}. We collected over 20,000 tactile frames for 15 individuals performing 4 types of dynamic hand gestures. Utilising this tactile information, we designed and implemented a gesture classifier equipped with an adaptive error correction mechanism. The gesture recognition system is characterised by its compact size ($<$ 5 MB), rapid speed ($<$ 1 ms), and minimal power requirements, making it appropriate for use in the majority of embedded systems including the one in \textit{etee} with the power consumption of 0.85 W. 

The rest of the paper is organised as follows. In Section \ref{sec:Hardware} we introduce our hardware and its architecture and describe hand gestures used in the dataset. Section \ref{sec:Dataset} outlines the data collection and preprocessing protocols and methods, Section \ref{sec:Base} explains two base multi-classification models used to distinguish dynamic gestures in the dataset: K-nearest neighbour (KNN) and Variational Autoencoder (VAE). Section \ref{sec:tuner} presents our adaptive error corrector designed for the task. Section \ref{sec:results}  summarises results and their brief discussions, and Section \ref{sec:conclusion} concludes the paper.

\section{Hardware}\label{sec:Hardware}
 
\begin{figure}[t]
        \centering
        \centerline{\includegraphics[width=0.48\textwidth]{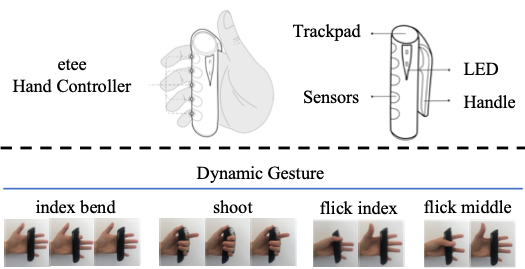}}
        \caption{\textbf{Top}: The wearable {\it etee} hand controllers used for collecting gesture recognition data in this work. The controller contains a trackpad, 5 finger sensors, LED and etc. \textbf{Bottom}: The names and movements of the 4 dynamic gestures collected for this study.}
        \label{fig:hardware}
    \end{figure}

  The controller hardware comprises of a capacitive touch sensor fusion unit, a printed circuit board (PCB) designed for a micro-controller unit (ESP32) and the support components like LED, battery and etc. The detailed hardware specifications can be found in the \textit{etee} website \cite{etee}. As shown in Fig.~\ref{fig:hardware} (top), the black sensors capture the finger proximity, touch, pressure signals across the hand-holding area. Each sensor covers a finger area and reflects the capacitive signal change triggered by the finger movement. The silicon shell works as an insulation layer to stabilise the sensor’s signal from the skin touch and regulate the signal strength for finger press. 

\section{Dataset}\label{sec:Dataset}
    
    \subsection{Data acquisition}
\begin{figure}[t]
        \centerline{\includegraphics[width=0.48\textwidth]{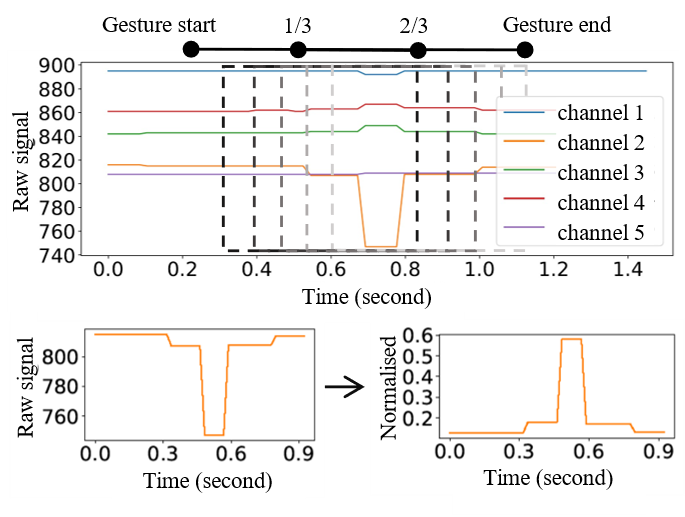}}
        \caption{\textbf{Top}: During data collection, gesture start and end were marked. A 500 ms time window, represented by the dashed box, was applied to the signal to extract segments, sliding from the black box to the grey box with each signal frame. \textbf{Bottom}: Original signals on the left are normalised between 0 and 1. The normalisation range is defined by each user and sensor, with zero meaning that the fingers are fully open and one indicating that the sensor is being applied with full pressure by the hand.}
        \label{fig:normalise}
    \end{figure}

    This section explains the data collection and preprocessing methods used in this article for the gesture recognition classification tasks. Fifteen individuals were recruited for the data collection, with 4 categories of hand gestures as shown in (see Fig.~\ref{fig:hardware}, bottom), namely "index bend", "shoot", "flick index", and "flick middle". These gestures are a sequence of movements depicted in the figure. Each recording consisted of 5 time-series signals from 5 sensors, with each sensor responsive to the movement of the corresponding finger position. Fig.~\ref{fig:normalise} shows the signal change during one hand gesture, "index bend", where the index finger starts to bend and then releases back to a straight position, resulting in an obvious variation in the signal at channel 2 (index finger position). The other sensors remain relatively constant. The dataset was extracted from the recordings in two ways: one is the exact dataset, with each sample marking a full gesture signal from start to end (as shown in Fig.~\ref{fig:normalise}), and the other is the sliding dataset, where a continuous 500 ms segment of the signal was extracted from 2/3 of the gesture until its end (indicated by the dashed box in Fig.~\ref{fig:normalise}). There are signals recorded from 4 dynamic gestures during the data collection, along with a "none" label to indicate when a user is not performing these gestures. The "none" gesture could be a random hand movement, resulting in a signal that does not fit a gesture pattern. 
    On the bottom, the original signals on the left were normalised between 0 and 1, with the normalisation range defined by each user and sensor. Zero indicates that the fingers are fully open, while one indicates that the sensor is applied with full pressure from the hand.
    
    The dataset was transformed into a 2D format by concatenating the 5 spectra from 5 sensors at each time step, resulting in a sample matrix with 5 channels and $x$ time stamps. Following a thorough analysis, it was determined that the average duration of a full gesture was approximately 500 ms. Based on this information, we then set $x$ to 20 when the transmitting frequency of the hardware was about 40 Hz. To align with the identified gesture duration, the time window in the sliding dataset was also established at 500 ms, ensuring consistent time stamps in each sample. The dataset was then normalised to the range [0,1] based on the maximum and minimum signals triggered by each user. Finally, the dataset was flattened to 100 features, with the first 20 features representing the thumb signal and the second 20 features representing the index signal.

    The signal patterns of all gestures vary from individual to individual due to different hand sizes and movement habits. As a result, the dataset is grouped by user and contains signals from different individuals. It is divided into four sets: training set, validation set, test set, and hold set, with the distribution of 8 users for training, 3 for validation, 2 for testing, and 2 for hold. The training set contains approximately 65\% of the samples, while the validation set contains about 15\%. The test set and hold set each contain 10\%. The hold set consists of samples that remain unchanged across the different sets.

    \subsection{Dimension Reduction}
\begin{figure}[t]
        \centerline{\includegraphics[width=0.48\textwidth]{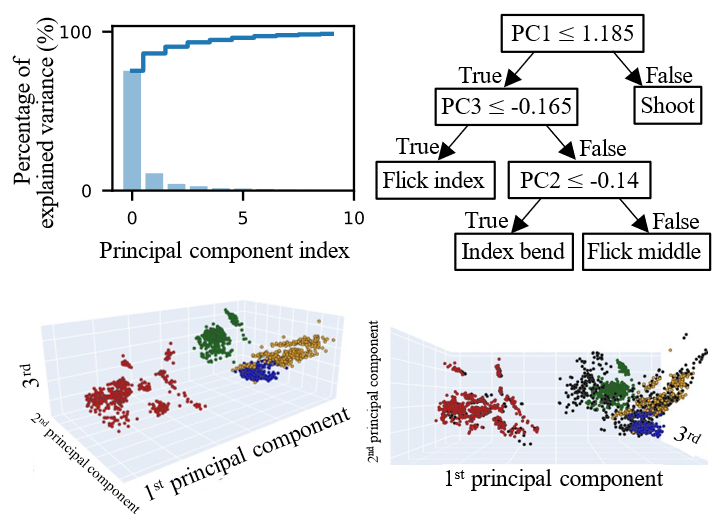}}
        \caption{\textbf{Top left}: The solid line is the accumulated percentage of explained variance and the bar is the individual percentage. The first three PCs cover over 95\% of the explained variance. \textbf{Top right}: All 100 principle components were fed to a decision tree for classification of 4 main dynamic gestures. Only three, PC1, PC2 and PC3 were required for the decision making. \textbf{Bottom left}: Three features - the top three PCs -  were used to visualise the dataset and show great sparsity among four gesture labels (colour represents the gesture here). \textbf{Bottom right}: An extra "none" label (black) were added in the dataset.}
        \label{fig:pca_analysis}
    \end{figure}

    Here, we employed principal component analysis (PCA) as a dimension reduction method. PCA is a commonly used technique in data mining and machine learning that identifies the directions in data that capture the most variance, and projects the data onto these directions \cite{Jolliffe2016}. This results in a set of orthogonal dimensions, called principal components (PCs), ranked by the variance they capture. The process is achieved via eigen-decomposition of the covariance matrix, and we used the built-in single value decomposition in Scikit-learn \cite{scikit} without centring the data as it has already been normalised, see Fig.~\ref{fig:normalise}. We selected the PCs based on the percentage of explained variance and a decision tree, confirming the importance of PC choices for classification. The first three PCs captured over 95\% of the explained variance of the original 100 features (as seen in Fig.~\ref{fig:pca_analysis}). A decision tree-based classification of the 100 transformed PCs also chose the first three PCs as the best splitters. Hence, these three features were selected as the reduced features for further classification analysis. 

\section{Base Model}\label{sec:Base}    
   
    Based on the three PC distribution of the dataset (Fig. \ref{fig:pca_analysis}), we noted a clear separation between gesture labels, with "none" labelled data enveloping other gestures. Given the limited number of training data to memorise on an edge device, we selected KNN as one of the base models due to its compatibility with the distribution pattern. To further evaluate model performance, we also employed a supervised VAE as a benchmark.
    
    \subsection{K-Nearest Neighbour}
    KNN is a simple and effective method for classification and regression \cite{Cover1967NearestClassification}. It is based on the idea that the input data can be classified or predicted based on the class or label of the nearest data points. The label of the output variable for a given input is determined by the K nearest data points, with $K$ being a user-defined parameter. KNN has been successfully applied in several domains, including medical health classification and fall prediction using foot sensors \cite{Xing2020knn, Liang2015knn}.
    
    However, KNN can be computationally expensive because all the distances between data points need to be computed and stored in order to find the nearest neighbours. Additionally, the accuracy of KNN predictions can decrease with an increasing number of features, due to the curse of dimensionality. This occurs when the number of dimensions increases, causing data points to become closer in the high-dimensional space, making it difficult for the KNN algorithm to separate them based on distance and make accurate predictions.
    
    The mathematical expression for KNN can be expressed as follows: let $x_i \in \mathbb{R}^d$ represent the $i$-th input data point and $y_i \in \mathbb{R}$ be the corresponding output variable. The goal of KNN is to predict the value of $\hat{y}$ for a new input $x$ based on the values of $y$ for the K nearest neighbours of $x$:
    
    $$\hat{y} = \frac{1}{K} \sum_{i \in \mathcal{N}_K(x)} y_i$$
    
    where $\mathcal{N}_K(x)$ is the set of indices of the K nearest neighbours of $x$.

    \subsection{Variational Autoencoder}
    This paragraph describes the VAE, a deep learning model used for representation learning and unsupervised learning \cite{Pinheiro2021VAE}. VAE comprises an encoder and a decoder, which map the input data to a lower-dimensional latent space and back to the original data space, respectively. A NN is used to learn the latent representation, where a prior distribution over the latent space is assumed and the encoder is trained to approximate the posterior distribution of the latent variables given the input data. This results in a flexible, scalable and interpretable representation of the data. Three latent representations are defined with likelihood functions in the form of Gaussian distributions.

    We define the latent feature $z$ and the observed data $x$. The posterior distribution is approximated in the encoder as $q_\phi(z|x)$. New samples are generated from the latent sample through a likelihood function, $p_\theta(x|z)$ in the decoder. The goal of VAE is to learn the parameters, $\phi$ and $\theta$, of the generative model that capture the underlying structure of the data. Hence we maximise the log-likelihood of the data coming out the decoder, which is bounded as:

    $$\log p_\theta(x) \ge -\mathrm{KL}(q_\phi(z|x) || p(z)) + \mathbb{E}{q_\phi(z|x)}\left[\log p_\theta(x|z)\right]$$
    
    where $\mathrm{KL}(\cdot | \cdot)$ is the Kullback-Leibler divergence and the right hand side of the above equation is called the Evidence Lower Bound (ELBO). Since maximising the $p_\theta(x)$ is typically computationally intractable, the object of VAE becomes maximising the ELBO instead, resulting a loss function as:
    
    $$\mathcal{L}(\theta, \phi) = \mathrm{KL}(q_\phi(z|x) || p(z)) - \mathbb{E}{q_\phi(z|x)}\left[\log p_\theta(x|z)\right]$$
    
    VAE is trained by minimising the loss function with respect to the parameters $\theta$ and $\phi$, resulting in a compact, continuous and interpretable latent representation that can be used for downstream tasks such as generation and reconstruction. The latent features are then fed to a two-layer neural network to predict the result. They were also used as reduced features for other classification models.
 
    \begin{figure}[t]
        \centerline{\includegraphics[width=0.48\textwidth]{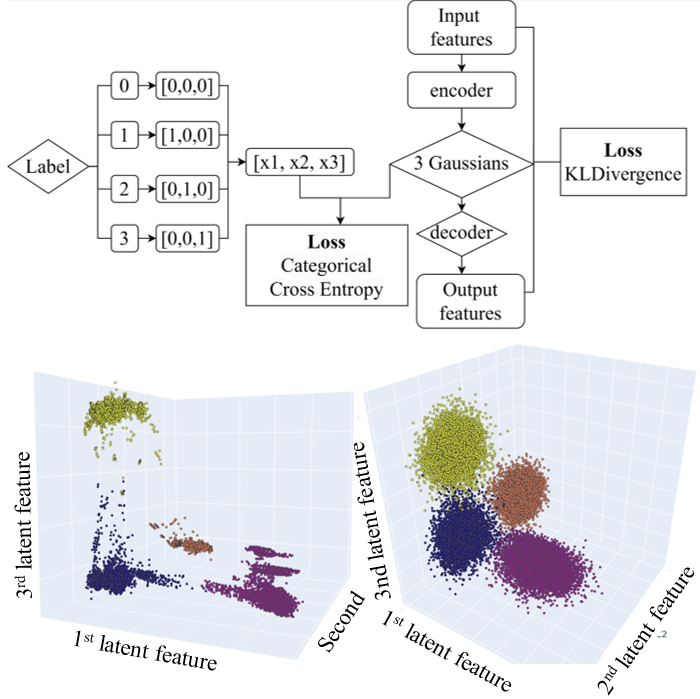}}
        \caption{\textbf{Top}: The structure of supervised VAE. \textbf{Bottom left}: the mean value of 3 latent features in the VAE. \textbf{Bottom right}: The randomly picked features from the Gaussian distribution of the three latent features.}
        \label{fig:VAE_3Dplot}
    \end{figure}
        
\section{Data Adaptive Tuner}\label{sec:tuner}

    The base models, the combination of PCA + KNN, the encoder + KNN and the supervised VAE, have an average accuracy of around 80\% (in Section \ref{sec:results}). However, they still make mistakes during the classification, leading to an unpleasant user experience. While increasing the amount of data from more users would alleviate some errors, it would also introduce others. To address this, we have introduced an error corrector into our legacy system, which corrects these errors without compromising the performance of the base model.

    The error corrector is based on the concentration of measure and the stochastic separation theorem in high dimensions \cite{error_corrector_2018}. According to the classical concentration of measure theory, i.i.d. random points in high dimensions are distributed in a thin layer of the sphere's surface.  Strikingly, with high probability, these points are also Fisher-separable from one another \cite{stochastic2017, stochastic2021}. This means that errors are linearly separable from the rest of the samples when the dimension is high, as has been demonstrated in various applications such as performance monitoring of computer numerical control milling processes and edge-based object detection \cite{corrector_application2021}.
    
    We evaluated two corrector algorithms to compare their performance combined with data from different high-dimension projections. These algorithms are the linear discriminant classifier \cite{corrector_fisher} and the centroid classifier.

    \subsection{Feature Analysis}
    
        For the classification of errors in high-dimensional space, the dataset is first projected into a high-dimensional feature space. There are several kernel functions that can perform this task, including the polynomial kernel, radial basis function (RBF) kernel, and nearest neighbour kernel through KNN.

        In our experiments, we used PCA, polynomial, and KNN kernels to transform the original 100 features of the normalised signal into a high dimension. The RBF kernel was not used, as the number of features generated by this kernel is much larger than the number of samples in the dataset, making it computationally expensive and difficult to predict.
        
        \begin{itemize}
            \item \textbf{PCA kernel:} We normalised the data in the selected feature space. After that, the dataset went through another PCA transformation and a whitening coordinate transformation to ensure the covariance matrix of the transformed data is an identity matrix. The features were then ranked from high to low by the percentage of explained variance. The first few PCs were extracted as features, with the number of features $\mathcal{N}_{pc}$ being a parameter. We tested this kernel with $\mathcal{N}_{pc}$ from 3 to 100.
            
            \item \textbf{Polynomial kernel:} All data was processed firstly using the PCA kernel with $\mathcal{N}_{pc}$. We then added nonlinear features through polynomial transformation with degree $\mathcal{N}_{poly}$. After that, we normalised the dataset and transform it through PCA and whitening process. In this kernel, the $\mathcal{N}_{pc}$ were tested from 2 to 20 and $\mathcal{N}_{poly}$ from 2 to 7.
            
            \item \textbf{KNN kernel:} All data were first processed using the PCA kernel with $\mathcal{N}_{pc}$. Next, we trained a KNN classifier based on the PCs. The $K_{nn}$ nearest neighbours of each sample were extracted as the new features. They were then normalised and transformed through PCA and whitening process again. In this kernel, The $K_{nn}$ was tested from 2 to 300 with a step of 10. 
        \end{itemize}

        \paragraph{Intrinsic Dimension}
        
            The intrinsic dimension is a significant concept in modern machine learning and has been widely discussed. Despite the number of features in a dataset, it does not always reflect its dimensionality \cite{stochastic2021}. For example, a dataset with three features distributed on a 2D plane has a dimension of 2 instead of 3. Various definitions of intrinsic dimension have been proposed, and here we use the Fisher separability statistic-based dimensionality proposed by Gorban and Tyukin \cite{error_corrector_2018}. This definition is consistent with PCA-based intrinsic dimension measurement and enables capturing low-dimensional fractal and fine-grain structures in data. 
            
\begin{figure*}
                    \centerline{\includegraphics[width=0.8\textwidth]{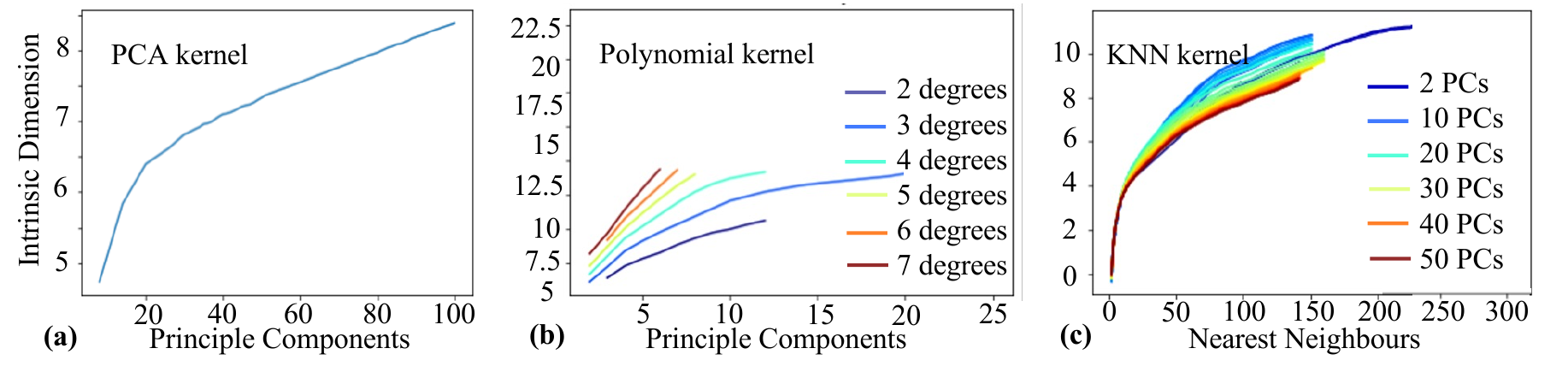}}
                    \caption{Intrinsic dimension of features from PCA kernel, Polynomial kernel and KNN kernel.}
                    \label{fig:intrinsic_dimension}
            \end{figure*}

            When the number of features in our dataset exceeds a certain limit, the intrinsic dimension cannot be accurately calculated due to the limited number of samples. However, as shown in Fig.~\ref{fig:intrinsic_dimension}, the trend of the intrinsic dimension can still be observed as we increase the value of certain parameters. The highest measurable intrinsic dimension observed does not exceed 12, which is roughly equal to the intrinsic dimension of the original inputs. To increase the dimensionality of the dataset, we combined features from two different kernels as a new feature space. These sets of features were applied to the following classifiers for error group classification and error correction. 
            
    \subsection{Error Correction}
        As five gesture labels were used in the base model (four dynamic gestures plus one "none" gesture), there are twenty possible error combinations, e.g. "shoot" predicted as "index bend". As depicted in Fig.~\ref{fig:pca_analysis}, some of the samples from different dynamic gesture labels in the training dataset have large enough gaps in the three PC spaces, making them easily separable from each other, while others are difficult to distinguish from one another. In Section\ref{subsec:ec}, we identify and discuss ten types of errors that occur after the base model prediction. These errors are predicted using an error group classifier. For each error group, we assign a new binary label that indicates whether a sample falls into a specific error pattern. For example, if the ground truth is "index bend" but the base model predicts it as "flick index," then the sample is assigned a positive label of 1. Conversely, if the sample does not fit this pattern, it is assigned a negative label of 0. The dataset was first separated into different error groups and then assigned new binary labels based on these patterns. These new subsets of the dataset were then used to train the error correctors, which are binary classifiers.

    \subsubsection{Binary Classifier}
        To perform binary classification from high-dimensional feature space, we tested two classifiers: linear discriminant analysis and centroid classifier.
        
        \paragraph{Linear Discriminant Analysis }
            Linear Discriminant Analysis (LDA) is a dimensionality reduction technique, commonly used in pattern recognition and machine learning \cite{Fisher1936}. It aims to find a linear combination of features that maximises the separation between classes. This is formalised as an optimisation problem that seeks a projection matrix $W$, maximising the ratio of between-class variance to within-class variance:

            $$J(W) = \frac{\mathrm{tr}(S_B W^T W)}{\mathrm{tr}(S_W W^T W)}$$
            
            where $S_B$ and $S_W$ are the between-class and within-class scatter matrices, respectively. The solution of this optimization problem is the projection matrix $W$, which gives the projected data by multiplying the original data matrix by $W$. LDA is particularly useful when the number of features is larger than the number of samples, as it can reduce the dimensionality while preserving class separability.

        \paragraph{Centroid Classifier}
            The paragraph describes the centroid-based classification method for supervised learning, which is a simple and effective approach. It is based on assigning a new data point to the class with the closest centroid, calculated as the average of the data points in the class. This instance-based learning algorithm uses the training data directly to make predictions, making it fast and easy to implement \cite{Trevor_book}. It can handle high-dimensional data and a large dataset.  The mathematical formulation is presented, including the definition of the centroid and the Euclidean distance between a data point, and the classification procedure of a new data point.
            
            Let $X = {x_1, \dots, x_n} \in \mathbb{R}^{n \times d}$ be a dataset containing $n$ samples and $d$ features, and let $y \in {0, 1, \dots, C}$ be the vector of class labels, where $C$ is the number of classes. The centroid of a class $c$ is defined as the average of the data points in the class:
            
            $$\mu_c = \frac{1}{n_c} \sum_{i: y_i = c} x_i$$
            
            where $n_c$ is the number of data points in the class. The distance between a data point $x$ and a centroid $\mu$ is typically measured using the Euclidean distance:
            
            $$d(x, \mu) = \sqrt{\sum_{j=1}^d (x_j - \mu_j)^2}$$
            
            To classify a new data point $x$, the centroid-based classifier finds the class with the closest centroid:
            
            $$\hat{y} = \arg\min_{c} d(x, \mu_c)$$

        \subsubsection{Receiver Operating Characteristic Curve}
        The Receiver Operating Characteristic (ROC) curve is a tool for evaluating binary classification models \cite{Fawcett2006}. It plots the True Positive Rate (TPR) against the False Positive Rate (FPR) at varying decision thresholds. This provides a visualisation of the trade-off between sensitivity and specificity in a binary classifier. In our case, the dataset can be imbalanced, causing TPR and FPR to be misleading. Therefore, we used the number of True Positive (TP) samples and False Positive (FP) samples to visualise the performance of binary classification models.
    

\section{Results and Discussion}\label{sec:results}


    \subsection{Base Model Performance}\label{subsec:accuracy}
        As data from the four gesture labels show clearer separation among each other, and the "none" gesture data is close to all the other four labels, we first analysed base models on the dataset consisting of the four gesture labels. Then, the chosen model structure was applied to the exact dataset with five labels, including the "none". Errors were observed in the 3D plots shown in Fig.~\ref{fig:intrinsic_dimension}, and these errors will be analysed in the next section.
        
        In this study, three base model systems were evaluated. Each system is a combination of dimension reduction and classification. The first system is a supervised VAE with three latent features, as depicted in Fig.~\ref{fig:VAE_3Dplot}. This system employed the losses: a loss from ELBO, and a categorical cross entropy between the latent features and transformed labels. The second base model IS dimension reduction using encoder part from VAE and classification using KNN. The third option is PCA + KNN. 
        
        To fairly estimate the accuracy of these models, k-fold cross validation was used. This technique assesses how the results of a machine learning algorithm generalise to an independent dataset. In this k-fold cross-validation, the ratio of users in the training, validation, test, and hold datasets remained 8:3:2:2, with users randomly allocated to different sets. Nearly 400 different combinations of users in the train, validation, and test datasets were performed. The hold dataset contained data from the same users across all combinations. In the supervised VAE model, only the training dataset was used for training. In the next two models, the training dataset was used for dimension reduction, and the validation set was used to train the KNN model. 

\begin{figure}
                \centerline{\includegraphics[width=0.5\textwidth]{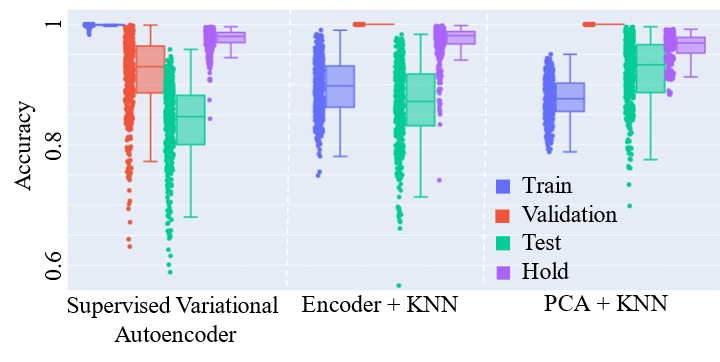}}
                \caption{The accuracy of $k$-fold cross-validation for three base models. The supervised VAE model, the encoder and PCA were trained on the train set. The KNN was trained on the validation set.}
                \label{fig:base_model_accuracy}
            \end{figure}

        Fig.~\ref{fig:base_model_accuracy} displays the accuracy calculated via $k$-fold cross-validation and three base model options. Each dot represents an accuracy computed from each dataset. In the test dataset, the PCA + KNN option exhibits the highest accuracy mean and the smallest variation among all three base models. The encoder + KNN option comes in second. In the hold dataset, where every dot has the same data, the accuracy means are similar across all options, but the smallest variation is observed in the PCA + KNN option. 

        \begin{table}[t]
        \caption{Computing Time (ms) on Intel i9 processor}
        \begin{center}
        \begin{tabular}{ccc}
        \textbf{\textit{Supervised VAE}}& \textbf{\textit{Encoder + KNN}}& \textbf{\textit{PCA + KNN}} \\
        \hline
        $>$ 20 & $>$ 20 & $<$ 1  \\
        \end{tabular}
        \label{tab1}
        \end{center}
        \end{table}

        We calculated the computing time of each model on a laptop with Intel i9 processor (shown in Table \ref{tab1}). A NN based model (supervised VAE and Encoder) takes significant time compared with PCA + KNN combination. As an embedded system generally has less computing power than a laptop, we chose PCA + KNN as the base model for the classification of 5 labels (4 dynamic gestures and 1 "none").

    \subsection{Error Corrector Performance}
    \label{subsec:ec}

        In the dataset that contains all 5 labels, we analysed the types of errors produced by the PCA + KNN base model. The base model was able to distinguish most of the gestures from each other, but struggled with differentiating "Flick Index"/"Index Bend" as well as "none"/others. After examining the base model predictions, we identified 10 possible error groups. To classify these error groups, we used an LDA classifier. After examining high-dimensional kernels with over 1000 parameter combinations, we selected PCA kernel with the first 9 PCs for the error group classification.

        In each error group, we trained a binary classifier to separate the errors from the rest of the data in that group. The training data came from the sliding data's train part. They were first projected into the high-dimensional feature space and then went through the binary classifier. We recorded ROC curves for the train and test dataset and  plotted the number of TPs against the number of FPs for each group. To accurately distinguish the errors without compromising the accuracy of the base model, we chose a high-dimensional kernel, a binary classifier, and a decision threshold for each group.
        
        By picking the right decision threshold, the classifier was able to achieve the maximum TPs with 0 FP. This allowed us to develop a corrector that only corrected the errors without affecting the base model's predictions.

\begin{figure}
                \centerline{\includegraphics[width=0.5\textwidth]{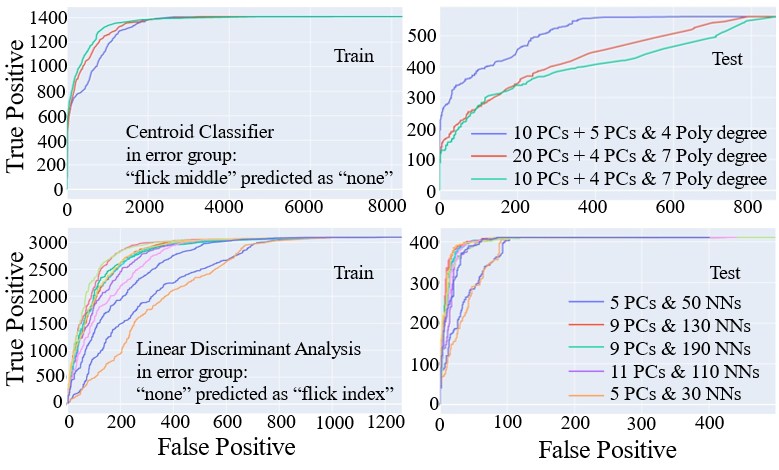}}
                \caption{\textbf{top}: The ROC curve calculated from 3 feature kernels in the centroid classifier for one error corrector. \textbf{bottom}: The ROC curve calculated from 11 feature kernels in LDA in another error group. Only 5 feature kernels are labelled here.}
                \label{fig:roc_curve}
            \end{figure}

        Here, we demonstrated some kernels with the best TPs at 0 FP in two error groups, as shown in Fig.~\ref{fig:roc_curve}. In one error group, where the ground truth is "flick middle" but the prediction is "none" in the base model (top figures in Fig.~\ref{fig:roc_curve}), the ROC curves were calculated from three different kernels using a centroid classifier in both the train and test dataset. The three feature kernels are: 1. The first 10 PCs from the PCA kernel and features from a polynomial kernel with $\mathcal{N}_{pc} = 5$ and $\mathcal{N}_{poly} = 4$; 2. 20 PCs from the PCA kernel plus a polynomial kernel of 5 PCs and 4 degrees; 3. 20 PCs from the PCA kernel plus a polynomial kernel of 5 PCs and 4 degrees. The ROC curves calculated from these three kernels show a reasonable TP number when the FP number is 0 in both the train and test datasets. When the decision threshold was defined as 0.955 for positive prediction, the centroid classifier detected 390 errors in the train set (1408 errors in 9796 samples from the base model) and 38 errors in the test set (562 errors in 1427 samples from the base model), thus ensuring error detection without damaging the base model prediction in both the train and test sets. In the bottom figures in Fig.~\ref{fig:roc_curve}, we demonstrate the ROC curves in an error group where the ground truth is "none" but predicted as "flick index" in the base model. There are ROC curves from 11 feature kernels in both the train and test sets, and we labelled 5 of them. In the LDA classifier for this error group, we were unable to define a threshold that detects errors without compromising the base model prediction. Hence, this model was not available in this error corrector system. If neither of these binary classifiers guarantees a positive TP number at 0 FP, we skip the implementation of an error corrector in this error group.

        After analysing the feature kernels and classifiers for 10 error groups, we selected the best combination of error correctors and evaluated their performance on a new dataset, as shown in Table~\ref{tab:accuracy compare}. The final prediction made by the combination of the error corrector and base model involves the following steps: 1) Make a prediction from the base model; 2) classify the sample into an error group; 3) if the base model's prediction fits the pattern of the error group, the sample is fed to the corresponding error corrector in that group; 4) if the error corrector identifies an error, the final result is corrected based on the pattern.

        \begin{table*}[t]
        \caption{The base model accuracy compared with corrector implantation}
        
        \centering
        \begin{tabular}{llcccccc}
        \hline
        \multicolumn{2}{l}{\multirow{2}{*}{}}         & \multicolumn{6}{c}{Accuracy}                                                                  \\ \cline{3-8} 
        \multicolumn{2}{l}{}                          & \multicolumn{2}{c}{Dataset 1} & \multicolumn{2}{c}{Dataset 2} & \multicolumn{2}{c}{Dataset 3} \\
        \multicolumn{2}{c}{Model }                     & base     & base + corrector   & base     & base + corrector   & base     & base + corrector   \\ \hline
        \multicolumn{2}{c}{Full dataset}              & 0.740    & 0.749              & 0.672    & 0.684              & 0.750    & 0.759              \\ \cline{1-2}
        \multirow{7}{*}{Error group} & group 2  & 0.618    & 0.618              & 0.453    & 0.459              & 0.463    & 0.490              \\
                                     & group 14 & 0.993    & 0.993              & 0.951    & 0.962              & 0.847    & 0.857              \\
                                     & group 15 & 0.702    & 0.702              & 0.930    & 0.907              & 0.913    & 0.918              \\
                                     & group 16 & 0.202    & 0.310              & 0.083    & 0.333              & 0.652    & 0.725              \\
                                     & group 17 & 0.829    & 0.829              & 0.913    & 0.870              & 0.997    & 1.000              \\
                                     & group 18 & 0.094    & 0.336              & 0        & 0.483              & 0.903    & 0.919              \\
                                     & group 19 & 0.273    & 0.273              & 0.430    & 0.430              & 0.646    & 0.646              \\ \hline
        \end{tabular}
        \label{tab:accuracy compare}
        \end{table*}

        Seven error groups were equipped with error correctors, as the rest of the groups could not be assigned a functional corrector. In Table~\ref{tab:accuracy compare}, we compared the accuracy of the base model to the accuracy of the base + corrector system across three datasets. The improvement in overall performance and performance in each error group suggests that the error corrector is an effective component of the gesture recognition system. Additionally, the computation time for this error correction mechanism for a single sample is less than 1 ms, making it feasible for use in embedded systems. The lightweight algorithm and short computation time enable this gesture recognition system to run on low-power devices.

\unskip

\section{Conclusions}\label{sec:conclusion}

    In this study, we presented a hand gesture recognition system that combines a simple base model and error correctors. The system uses low-dimensional capacitive sensor signals measured on the \textit{etee} hand controller, rather than high-dimensional image data, which is computationally expensive. Through PCA analysis and decision tree analysis, we extracted three features for base model classification. The performance of the PCA + KNN combination in the base model was compared to a benchmark of supervised VAE, and showed not only comparable accuracy, but also over 20 times faster computing time. To address occasional errors, we added an adaptive error correction mechanism to the system, where each error group is assigned a corresponding corrector. The results demonstrated an improvement in the overall accuracy, as well as performance gains with respect to the individual error groups. Given the small system's size and fast computation time, the adaptive error correction mechanism makes the gesture recognition system ideal for live useage in low-power (0.85 W) hand controllers like the \textit{etee}. Although our study provides a viable solution for customising gesture recognition systems on edge devices, it is important to acknowledge its limitations. While the error correction mechanism is capable of identifying errors in some groups, it may not be able to distinguish errors in others. Further research is needed to gain insights into the underlying reasons for this discrepancy and into determining alternative ways to deal with AI errors.

\vspace{6pt}







\bibliography{references}

\end{document}